%% file: neurips_2020.tex
\documentclass{article}

\usepackage[nonatbib, preprint]{neurips_2020}

\usepackage[utf8]{inputenc} % allow utf-8 input
\usepackage[T1]{fontenc}    % use 8-bit T1 fonts
\usepackage{hyperref}       % hyperlinks
\usepackage{url}            % simple URL typesetting
\usepackage{booktabs}       % professional-quality tables
\usepackage{amsfonts}       % blackboard math symbols
\usepackage{nicefrac}       % compact symbols for 1/2, etc.
\usepackage{microtype}      % microtypography
\usepackage{algorithm}
\usepackage{algorithmic}
\usepackage{multicol}
\usepackage{comment}
\usepackage{fixmath}
\usepackage{bm}
\usepackage{subcaption}
\usepackage{caption}
\usepackage{multirow}
\usepackage{makecell}
\usepackage{amsmath}
\usepackage{cleveref}

\crefformat{section}{\S#2#1#3} 
\crefformat{subsection}{\S#2#1#3}
\crefformat{subsubsection}{\S#2#1#3}

\makeatletter

\def \hfillx {\hspace*{-\textwidth} \hfill}

\usepackage{wrapfig}

\newcommand{\distas}[1]{\mathbin{\overset{#1}{\kern\z@\sim}}}%
\newsavebox{\mybox}\newsavebox{\mysim}
\newcommand{\distras}[1]{%
  \savebox{\mybox}{\hbox{\kern3pt$\scriptstyle#1$\kern3pt}}%
  \savebox{\mysim}{\hbox{$\sim$}}%
  \mathbin{\overset{#1}{\kern\z@\resizebox{\wd\mybox}{\ht\mysim}{$\sim$}}}%
}
\makeatother

\def\etal{\emph{et al.}}

\title{Unsupervised Video Decomposition using Spatio-temporal Iterative Inference}

\author{%
\hspace{-0.25in}    Polina Zablotskaia$^{1,2,3,}$\thanks{This work was done during an internship at Borealis AI. Correspondence to \texttt{pzablots@cs.ubc.ca}.}
\hspace{0.15in} Edoardo A. Dominici$^{1}$
\hspace{0.15in} Leonid Sigal$^{1,2,3,4}$
\hspace{0.15in} Andreas M. Lehrmann$^{2}$  \\
$^1$Department of Computer Science, University of British Columbia\\
$^2$Borealis AI, Vancouver, BC, Canada \\
$^3$Vector Institute for AI \hspace{1in}   $^4$CIFAR AI Chair \\ 
}

\usepackage[pdftex]{graphicx}
\usepackage[dvipsnames]{xcolor}
\begin{document}

\maketitle

\begin{abstract}
Unsupervised multi-object scene decomposition is a fast-emerging problem in representation learning. Despite significant progress in static scenes, such models are unable to leverage important dynamic cues present in video. We propose a novel spatio-temporal iterative inference framework that is powerful enough to jointly model complex multi-object representations and explicit temporal dependencies between latent variables across frames. This is achieved by leveraging 2D-LSTM, temporally conditioned inference and generation within the iterative amortized inference for posterior refinement. Our method improves the overall quality of decompositions, encodes information about the objects' dynamics, and can be used to predict trajectories of each object separately. Additionally, we show that our model has a high accuracy even without color information. We demonstrate the decomposition, segmentation, and prediction capabilities of our model and show that it outperforms the state-of-the-art on several benchmark datasets, one of which was curated for this work and will be made publicly available.

\end{abstract}

\input{intro.tex}
\input{related.tex}
\input{background.tex}
\input{method.tex}
\input{experiments.tex}
\input{conclusion}

\begin{ack}
This work was funded, in part, by the Vector Institute for AI, Canada CIFAR AI Chair, NSERC CRC and an NSERC DG and Discovery Accelerator Grants.
\end{ack}

%\section*{Broader Impact}

%\subsection*{Field of Research}
%We think our work makes considerable progress in the field of unsupervised learning, investigating how foundational principles of vision affect low-level processing tasks. This is a very important direction of research, as described in Section~\ref{sec:intro}, and we expect future work to build upon or further generalize our model.

%\subsection*{Society}
%We don't expect our work to be used as-is, but rather as a stepping stone for future improvements and more complex applications. We briefly discuss possible side effects arising from derived applications.
%Since unsupervised models reduce the effort and simplify the process of mining complex information, data mining applications need to be careful to respect current privacy regulations. 
%Recent events brought this issue to the public's attention \cite{granville_2018} \cite{kassner_2015}.
%More generally, learning models are vulnerable to systematic bias unless specifically addressed. We refer to a recent survey conducted by Mehrabi et al \cite{mehrabi2019survey} on research related to this issue.

\medskip

\small

% ---- Bibliography ----
%
% BibTeX users should specify bibliography style 'splncs04'.
% References will then be sorted and formatted in the correct style.
%
\bibliographystyle{splncs04}
\bibliography{egbib}

\input{supp.tex}

\end{document}

%% file: intro.tex
\section{Introduction}
\label{sec:intro}
\vspace{-0.07in}
Unsupervised representation learning, which has a long history dating back to Boltzman Machines \cite{Hinton1986} and original works of Marr \cite{Marr1970}, has recently emerged as one of the important directions of research, carrying the newfound promise of alleviating the need for excessively large and fully labeled datasets. % that are currently required by most neural architectures.
More traditional representation learning approaches focus on unsupervised ({\em e.g.}, %  denoising
autoencoder-based \cite{pathak2016cvpr,vincent2008icml}) or self-supervised \cite{noroozi2016eccv,vondrick2016cvpr,zhang2016eccv} learning of {\em holistic} representations that, for example, are tasked with producing (spatial \cite{noroozi2016eccv}, temporal \cite{vondrick2016cvpr}, or color \cite{zhang2016eccv}) %predictive
encodings of images or patches. The latest and most successful methods along these lines include ViLBERT \cite{Lu2019nips} and others \cite{sun2019iccv,tan2019emnlp} that utilize powerful transformer architectures \cite{vaswani2017attention} coupled with proxy multi-modal tasks ({\em e.g.}, masked token prediction or visua-lingual alignment). 
Learning of good {\em disentangled}, spatially {\em granular}, representations that are, for example, able to decouple object appearance and shape in complex visual scenes consisting of multiple moving objects remains elusive.

Recent works that attempt to address this challenge can be characterized as: (i) attention-based methods \cite{crawford2019spatially,eslami2016attend}, which infer latent representations for each object in a scene, and (ii) iterative refinement models \cite{greff2019multi,greff2017neural}, which decompose a scene into a collection of components by grouping pixels. % \andreas{Iterative refinements themselves, as in amortized itertive inference, do not incentivize a decomposition.} 
Importantly, the former have been limited to latent representations at object- or image patch-levels, while the latter class of models have illustrated the ability for more granular latent representations at the pixel (segmentation)-level. Specifically, most refinement models learn pixel-level generative models driven by spatial mixtures~\cite{greff2017neural} and utilize amortized iterative refinements~\cite{marino2018iterative} for inference of disentangled latent representations within the VAE framework \cite{kingma2014iclr}; a prime example is IODINE \cite{greff2019multi}. However, while providing a powerful model and abstraction which is able to segment and disentangle complex scenes, IODINE \cite{greff2019multi} and other similar architectures are fundamentally limited by the fact that they only consider images. Even when applied for inference in video, they process one frame at a time. This makes it excessively challenging to discover and represent individual instances of objects that may share properties such as appearance and shape but differ in dynamics. 
% \andreas{That is true, but do we have convincing experiments illustrating our model's ability to disambiguate based on temporal dynamics?}

\begin{figure}[t]
\centering
\includegraphics[width=13.4cm]{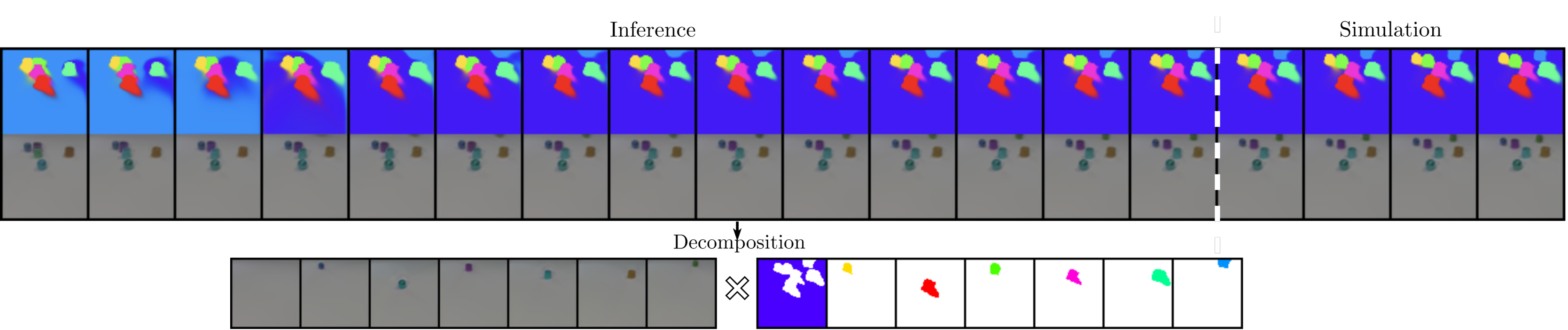} \\
\caption{{\bf Unsupervised Video Decomposition.} Our approach allows to infer precise segmentations of the objects via interpretable latent representations, that can be used to decompose each frame and simulate the future dynamics, all in unsupervised fashion. Whenever a new object emerges into a frame the model dynamical adapts and uses one of the segmentation slots to assign to the new object.}
\label{fig:teaser}
\vspace{-0.07in}
\end{figure}

In computer vision, it has been a long-held belief that motion carries important information for segmenting objects \cite{jepson2002eccv,weiss1996cvpr}. Armed with this intuition, we propose a spatio-temporal amortized inference model capable of not only unsupervised multi-object scene decomposition, but also of learning and leveraging the implicit probabilistic dynamics of each object from perspective  % complex \andreas{Overclaim?} 
raw video alone. This is achieved by introducing temporal dependencies between the latent variables across time. As such, IODINE~\cite{greff2019multi} could be considered a special (spatial) case of our spatio-temporal formulation. Modeling temporal dependencies among video frames also allows us to make use of conditional priors \cite{chung2015recurrent} for variational inference,
%which, as a consequence, leads
leading to more accurate and efficient inference results. The resulting model, illustrated in Fig.~\ref{fig:teaser}, achieves superior performance on complex multi-object
%public
benchmarks with respect to state-of-the-art models, including R-NEM~\cite{van2018relational} and IODINE~\cite{greff2019multi}.

{\bf Contributions.} We propose a new spatio-temporal amortized inference model that is not only capable of multi-object video decomposition in an unsupervised manner but also learns and models the probabilistic dynamics of each object from complex raw video data by leveraging temporal dependencies between the latent random variables at each frame. To the best of our knowledge this is the first spatio-temporal model of this kind. Our model has a number of appealing properties, including temporal extrapolation (prediction), computational efficiency, and the ability to work with complex data exhibiting non-linear dynamics, colors, and changing number of objects within the same video sequence (\emph{e.g.}, due to objects exiting and entering the scene). In addition, we introduce an entropy prior to improve our model’s performance in scenarios where object appearance alone is not sufficiently distinctive (\emph{e.g.}, greyscale data). Finally, we illustrate state-of-the-art performance on challenging multi-object benchmark datasets (Bouncing Balls and CLEVRER), outperforming results of R-NEM \cite{van2018relational} and IODINE \cite{greff2019multi} in terms of segmentation, prediction, and generalization.

% Marr, D (1970). A theory for cerebral neocortex. Proceedings of the Royal Society of London, Series B, 176, 161-234.

% Hinton, GE & Sejnowski, TJ (1986). Learning and relearning in Boltzmann machines. In DE Rumelhart, JL McClelland and the PDP research group, editors, Parallel Distributed Processing: Explorations in the Microstructure of Cognition. Volume 1: Foundations, Cambridge, MA: MIT Press, 282-317.

%% file: related.tex
\section{Related work}
\label{sec:related}

\vspace{-0.07in}
 \noindent \textbf{Unsupervised Scene Representation Learning.} Unsupervised scene representation learning has a rich history. Generally, these methods can be divided into two groups: attention-based methods, which infer latent representations for each object in a scene, and more complex and powerful iterative refinement models, which often make use of spatial mixtures and can decompose a scene into a collection of precisely estimated components by grouping pixels together. %Models from the first category, 
 {\em Attention-based} methods, such as AIR~\cite{eslami2016attend} and SPAIR~\cite{crawford2019spatially}, decompose scenes into latent variables representing the appearance, position, and size of the underlying objects. However, both methods can only infer the objects' bounding boxes (not segmentations) and have not been shown to work on non-trivial 3D scenes with perspective distortions and occlusions. MoNet~\cite{burgess2019monet} is the first model in this family tackling more complex data and infering representations that can be used for instance segmentation of the objects and individual reconstructions. On the other hand, it is not a probabilistic generative model and thus not suitable for density estimation. GENESIS~\cite{engelcke2019genesis} extends MoNet and alleviates some of its limitations by introducing a probabilistic framework and allowing for spatial relations between the objects. 
 {\em Iterative refinement} models started with 
 Tagger~\cite{greff2016tagger} 
 % falls into the second category above and 
 that explicitly reasons about the segmentation of its inputs and features. However, it does not allow explicit latent representations and cannot be scaled to larger and more complex images. NEM~\cite{greff2017neural}, as an extension of Tagger, uses a spatial mixture model inside an expectation maximization framework but is limited to binary data. Finally, IODINE~\cite{greff2019multi} is a notable example of a model employing iterative amortized inference w.r.t. a spatial mixture formulation and achieves state-of-the-art performance in scene decomposition and segmentation. Furthermore, it can cope with complex data, including occlusions, and uses an auxilliary component to separate the objects from the background. 
 % Unlike MoNet, it is also a proper probabilistic model and has a more natural ability to extend to sequential data. 
 %\andreas{Too much praise for the model closest to our formulation?} \leon{We can cut last sentence I think. Also, need to clarify the split between categories.} 
%In fact, IODINE has a preliminary experiments of  readily extending it to sequential data, but this kind of direct application of the static model to sequences doesn't allow explicit modelling of dynamics as well as it doesn't guarantee successful behaviour in case of newly emerging objects or complex motions. It also suffers from poor decomposition for several first frames and it hasn't been extended to future predictions.

\noindent \textbf{Unsupervised Video Tracking and Object Detection.} SQAIR ~\cite{kosiorek2018sequential}, SILOT~\cite{crawford2019exploiting} and  SCALOR~\cite{jiang2019scalor} are temporal extensions of the static attention-based models that are tailored to tracking and object detection tasks. SQAIR is restricted to binary data and % assumes a coarser granularity 
operates at the level of bounding boxes. SILOT and SCALOR are more expressive and can cope with cluttered scenes, a larger numbers of objects, and dynamic backgrounds but do not work on colored perspective\footnote{Perspective videos are more complex as objects can occlude one another and change in size over time.} data; accurate segmentation remains a challenge. % for them. 
% Finally, STOVE~\cite{kossen2019structured} is an attention-based model focusing on physics-driven learning and simulation on synthetic datasets.
Finally, STOVE~\cite{kossen2019structured} focuses on physics-driven learning and simulation.

\noindent \textbf{Unsupervised Video Decomposition and Segmentation.} Models employing spatial mixtures and iterative inference in a temporal setting are closest to our method from a technical perspective. Notably, there are only few models falling into this line of work: RTagger~\cite{premont2017recurrent} is a recurrent extension of Tagger but inherits the limitations of its predecessor. R-NEM~\cite{van2018relational} effectively learns the objects' dynamics and interactions through a relational module % and can produce segmentations 
but is limited to orthographic binary data. %~\cite{goel2018unsupervised} - very poor masks, focus on RL-\polina{cannot decide where to put it}.

\noindent \textbf{Non-representation Learning Methods.} Orthogonal to unsupervised representation learning for instance segmentation and object detection are methods relying on fully labeled data, including Mask R-CNN~\cite{he2017mask}, Yolo V3~\cite{redmon2018yolov3}, and Fast R-CNN~\cite{girshick2015fast}. Alternatively, hand-crafted features can be used, as demonstrated in~\cite{felzenszwalb2004efficient,shi2000normalized}. Unsupervised video segmentation also plays an important role in reinforcement learning: MOREL~\cite{goel2018unsupervised} takes an optical flow approach to segment the moving objects, while others use RL agents to infer  segmentations~\cite{casanova2020reinforced}.

%% file: background.tex
\section{Dynamic Video Decomposition}
\vspace{-0.07in}
We now introduce our dynamic model for unsupervised video decomposition. Our approach builds upon a generative model of multi-object representations~\cite{greff2019multi} and leverages elements of iterative amortized inference~\cite{marino2018iterative}. We briefly review both concepts~(\cref{sec:bg}) and then 
introduce our model~(\cref{sec:method}).
% We will now introduce our dynamic model for unsupervised scene decomposition. Our approach builds upon a generative model of multi-object representations~\cite{greff2019multi} and leverages elements of iterative amortized inference~\cite{marino2018iterative}. We briefly review both concepts~(Section~\ref{sec:bg}) and then % move on to a description of 
% introduce our model~(Section~\ref{sec:method}).
\subsection{Background}
\vspace{-0.07in}
\label{sec:bg}
\paragraph{Multi-Object Representations.}
The multi-object framework introduced in~\cite{greff2019multi} decomposes a static image ${\bf x}=(x_i)_i \in \mathbb{R}^D$ into $K$ objects (including background). Each object is represented by a latent vector ${\bf z}^{(k)} \in \mathbb{R}^M$ capturing the object's unique appearance and can be thought of as an encoding of common visual properties, such as color, shape, position, and size. For each ${\bf z}^{(k)}$ independently, a broadcast decoder~\cite{watters2019spatial} generates pixelwise pairs $(m_{i}^{(k)},\mu_{i}^{(k)})$ describing the assignment probability and appearance of pixel $i$ for object $k$. Together, they induce the generative image formation model
\begin{align}\label{eq:genModel}
p({\bf x}|{\bf z}) = \prod_{i=1}^{D}\sum_{k=1}^{K}m_{i}^{(k)}\mathcal{N}(x_i;\ \mu_{i}^{(k)}, \sigma^2),
\end{align}
where ${\bf z}=({\bf z}^{(k)})_k$, $\sum_{k=1}^{K} m_{i}^{(k)} = 1$ and $\sigma$ is the same and fixed for all $i$ and $k$. The original image pixels can be reconstructed from this probabilistic representation as $\widetilde{x}_i = \sum_{k=1}^{K}m_{i}^{(k)}\mu_{i}^{(k)}$.
\vspace{-0.07in}
\paragraph{Iterative Amortized Inference.}
Our approach leverages the iterative amortized inference framework~\cite{marino2018iterative}, which uses the learning to learn principle~\cite{Andrychowicz16} to close the amortization gap~\cite{Cremer17} typically observed in traditional variational inference. The need for such an iterative process arises due to the multi-modality of~Eq.\eqref{eq:genModel}, which results in an order invariance and assignment ambiguity in the approximate posterior that standard variational inference cannot overcome~\cite{greff2019multi}.

The idea of amortized iterative inference is to start with randomly guessed parameters $\bm{\lambda}_1^{(k)}$ for the approximate posterior $q_{\bm{\lambda}}({\bf z}_1^{(k)}|{\bf x})$ and update this initial estimate through a series of $R$ refinement steps. Each refinement step $r\in \{1,\ldots,R\}$ first samples a latent representation from $q_{\bm{\lambda}}$ to evaluate the ELBO $\mathcal{L}$ and then uses the approximate posterior gradients $\nabla_{\bm{\lambda}}\mathcal{L}$ to compute an additive update $f_\phi$, producing a new parameter estimate $\bm{\lambda}_{r+1}^{(k)}$:
\begin{align}
  {\bf z}_{r}^{(k)}\mkern9mu \distas{k}\mkern9mu  &q_{\bm{\lambda}}({\bf z}_{r}^{(k)}|{\bf x}),~~~~~~~~~~~~~~~~
  \bm{\lambda}_{r+1}^{(k)}\mkern9mu \xleftarrow{k}\mkern9mu  \bm{\lambda}_{r}^{(k)} + f_\phi({\bf a}^{(k)}, {\bf h}_{r-1}^{(k)}), \label{eq:iai}
\end{align}
where ${\bf a}^{(k)}$ is a function of ${\bf z}_{r}^{(k)}$, ${\bf x}$, $\nabla_{\bm{\lambda}}\mathcal{L}$, and additional inputs. The function $f_{\phi}$ consists of a sequence of convolutional layers and an LSTM. The memory unit takes as input a hidden state ${\bf h}_{r-1}^{(k)}$ from the previous refinement step.
%The full description of the inputs and the refinement network can be found in the IODINE paper\cite{greff2019multi}.

%% file: method.tex
\vspace{-0.02in}
\subsection{Spatio-Temporal Iterative Inference}
\label{sec:method}
\vspace{-0.07in}
Our proposed model builds upon the concepts introduced in the previous section and enables robust learning of dynamic scenes through spatio-temporal iterative inference. Specifically, we consider the task of decomposing a video sequence ${\bf x} = ({\bf x}_t)_{t=1}^T=(x_{t,i})_{t,i=1}^{T,D}$ into $K$ slot sequences $({\bf m}_t^{(k)})_t$ and $K$ appearance sequences $({\bm \mu}_t^{(k)})_t$. To this end, we introduce explicit temporal dependencies into the sequence of posterior refinements and show how to leverage this contextual information during decoding with a generative model. The resulting computation graph can be thought of as a 2D grid with time dimension $t$ and refinement dimension $r$~(Fig.~\ref{fig:overview_a}). Propagation of information along these two axes is achieved with a 2D-LSTM~\cite{graves2007multi}~(Fig.~\ref{fig:overview_b}), which allows us to model the joint probability over the entire video sequence inside the iterative amortized inference framework.
%Our novel iterative inference algorithm explicitly conditions the posterior on the previous refinement steps as well as on the previous video frames.
The proposed method is expressive enough to model the multimodality of our image formation process and posterior, yet its runtime complexity is smaller than that of its static counterpart.

 %In this section we discuss our contribution, spatio-temporal iterative inference. We first start with a variational objective and a motivation behind it, followed by a novel inference and generation procedures. Finally, in Section \ref{train} we describe the new entropy loss, training, and how the model is adapted to the simulation task.

\subsubsection{Variational Objective}
\vspace{-0.06in}
\label{vo}
Since exact likelihood training is intractable, we formulate our task in terms of a variational objective. In contrast to traditional optimization of the evidence lower bound (ELBO) through static encodings of the approximate posterior, we incorporate information from two dynamic axes: (1) variational estimates from previous refinement steps; (2) temporal information from previous frames. Together, they form the basis for spatio-temporal variational inference via iterative refinements.
%But we need to take into account that out method is based on the approximation of the posterior parameters through the sequence of refinement iterations. We also know that each of the video frames in dependent on the previous ones. To better model the variational posterior, prior and the likelihood we need to condition the distributions on all the previous latent representations and on the previous frames.
Specifically, we train our model by maximizing the following ELBO objective\footnote{For simplicity, we drop references to the object slot $\bullet^{(k)}$ from now on and formulate all equations on a per-slot basis.}:
\begin{equation}
\begin{aligned}\label{eq:obj}
  \mathcal{L}_{\textrm{ELBO}}({\bf x})  =  \mathbb{E}_{q_{\bm\lambda}({\bf z}_{\leq T,R}|{\bf x}_{\leq T})}\sum_{t=1}^{T}\sum_{r=1}^{\widehat{R}}\Bigl[&\beta \log\left(p\left({\bf x}_t|{\bf x}_{<t}, {\bf z}_{\leq t, r}\right)\right)  \\
  & - \textrm{KL}(q_{\bm\lambda}({\bf z}_{t,r}|{\bf x}_{\leq t}, {\bf z}_{< t,r})\ ||\ p({\bf z}_t|{\bf x}_{<t},{\bf z}_{<t}))\Bigr],
\end{aligned}
\end{equation}
where the first term expresses the reconstruction error of a single frame and the second term measures the divergence between the variational posterior and the prior. The relative weight between terms is controlled with a hyperparameter $\beta$~\cite{higgins2017beta}. Furthermore, to reduce the overall complexity of the model and to make it easier to train, we set $\widehat{R}:=\max(R-t, 1)$ (see Fig.~\ref{fig:overview} for an illustration). Compared to a static model, which infers each frame independently, reusing information from previous refinement steps also makes our model more computationally efficient.
In the next sections, we discuss the form of the conditional distributions in Eq.\eqref{eq:obj} in more detail.
\subsubsection{Inference and Generation}
\begin{figure*}[t!]
    \centering
    \begin{subfigure}[t]{0.5\textwidth}
        \centering
        \includegraphics[width=0.9\linewidth]{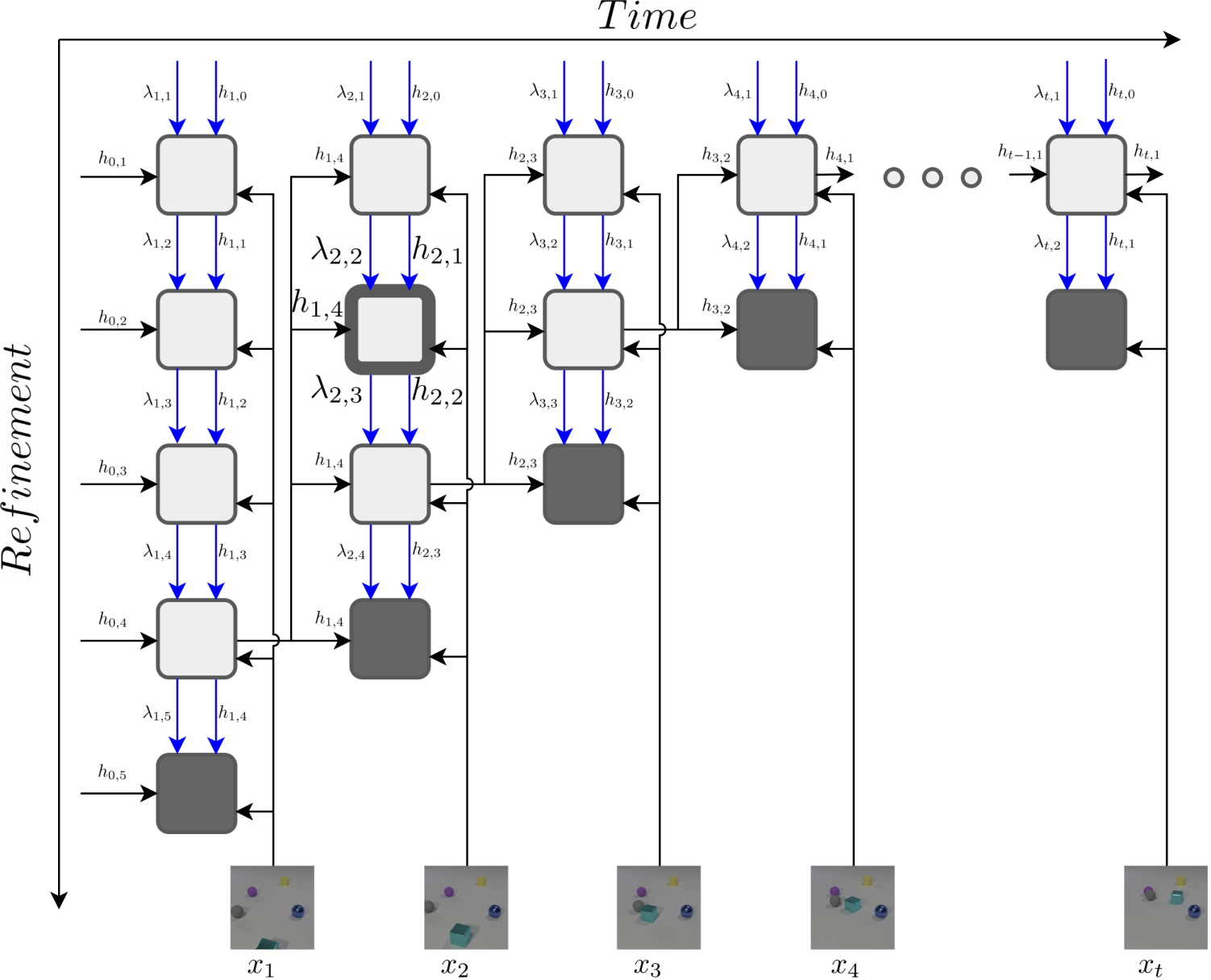}
        \caption{}
        \label{fig:overview_a}
    \end{subfigure}%
    ~
    \begin{subfigure}[t]{0.5\textwidth}
        \centering
        \includegraphics[width=0.9\linewidth]{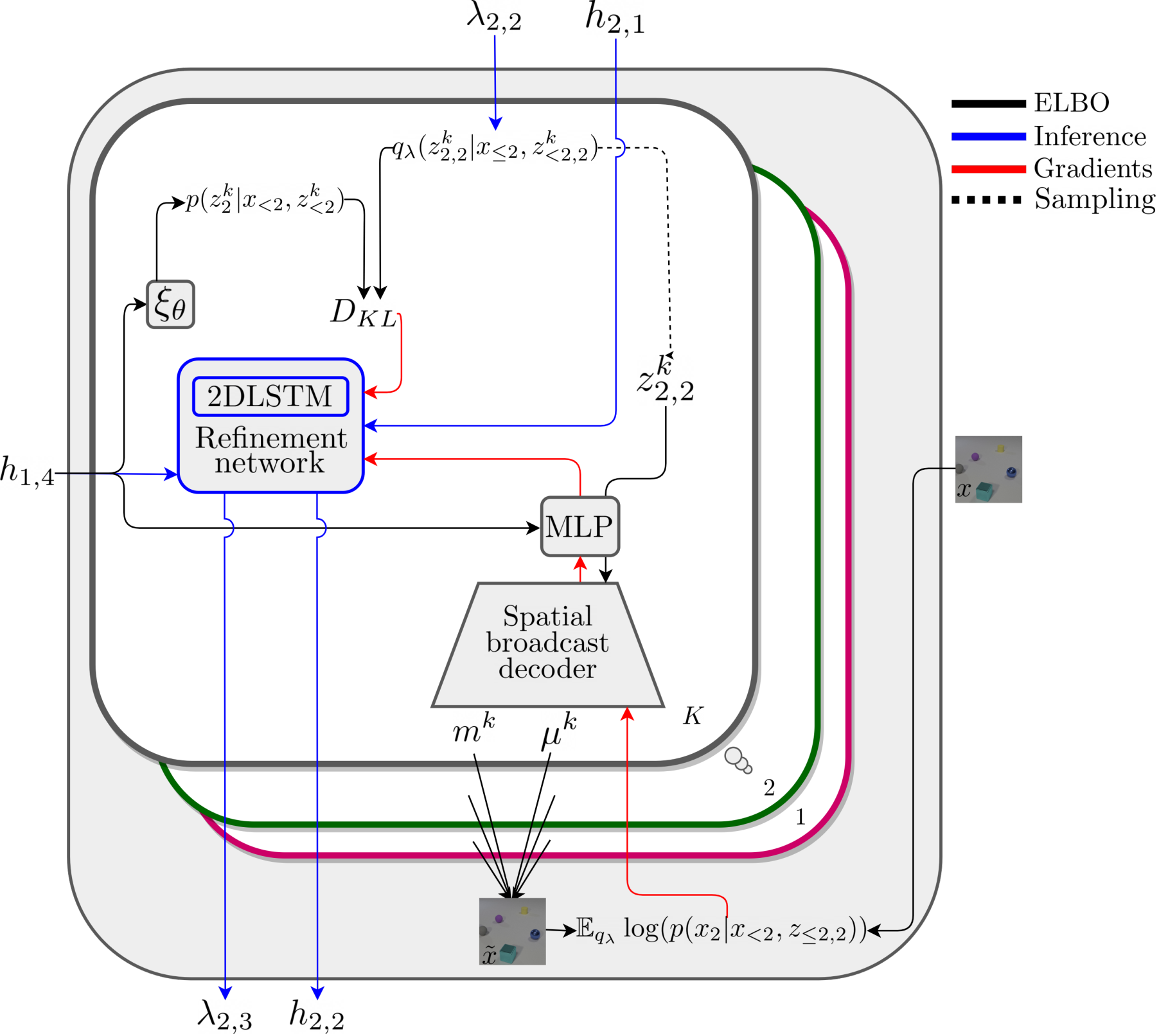}
        \caption{}
        \label{fig:overview_b}
    \end{subfigure}
    \caption{{\bf Model Overview.} ({\bf a}) Inference in our model passes through a 2D grid in which light gray cell $(r,t)$ represents the $r$-th refinment at time $t$, dark gray cells are where the final reconstruction is computed and no refinement is needed . Each light gray cell receives three inputs: a refinement hidden state ${\bf h}_{t,r-1}$, a temporal hidden state ${\bf h}_{t-1, \widehat{R}}$, and posterior parameters $\bm\lambda_{t,r}$. The outputs are a new hidden state ${\bf h}_{t,r}$ and new posterior parameters $\bm\lambda_{t,r+1}$. ({\bf b}) An example of the internal structure of the highlighted cell from Fig.~(a). We process the inputs with the help of a spatial broadcast decoder~\cite{greff2019multi} and a 2D LSTM~\cite{graves2007multi}. The rest of the light gray cells have the same structure.}
    \label{fig:overview}
    \vspace{-0.08in}
\end{figure*}
\label{ig}
\vspace{-0.06in}
\noindent \textbf{Posterior Refinement.} Optimizing Eq.\eqref{eq:obj} inside the iterative amortized inference framework~(Section~\ref{sec:bg}) requires careful thought about the nature and processing of the hidden states. While there is vast literature on the propagation of a single signal, including different types of RNNs~\cite{hochreiter1997long, cho2014learning, gravesblstm, chung2016hierarchical} and transformers~\cite{vaswani2017attention}, the optimal solution for multiple axes with different semantic meaning ({\em i.e.}, time and refinements) is less obvious.
%This setup doesn't require a modelling of hierarchically structured data or a bi-directional information flow.
%The uni-directional MD-LSTM~\cite{graves2007multi} is a type of model that can best satisfy properties of our data and the inference procedure.
Here, we propose to use a 2D version of the uni-directional MD-LSTM~\cite{graves2007multi} to compute our variational objective (Eq.\eqref{eq:obj}) in an iterative manner. In order to do so, we replace the traditional LSTM in the refinement network~(Eq.\eqref{eq:iai}) with a 2D extension. This extension allows the posterior gradients to flow through both the grid of the previous refinements and the previous time steps (see Fig.~\ref{fig:overview_a}). Writing ${\bf z}_{t,r}$ for the latent encoding at time $t$ and refinement $r$, we can formalize this new update scheme as follows:
%\polina{Do we need below some formalization of 2DLSTM, i.e. formulas for the hidden state, to empathises the impact of 2DLSTM. Please see the equations (1) and (2) here https://arxiv.org/pdf/1506.02216.pdf}
\begin{align}\label{eq:st-iai}
  {\bf z}_{t,r} \sim  q_{\bm\lambda}({\bf z}_{t,r}|{\bf x}_{\leq t}, {\bf z}_{< t,r}),~~~~~~~~~~~~~~~~
  {\bm\lambda}_{t,r+1} \leftarrow {\bm\lambda}_{t,r} + f_\phi({\bf a}, {\bf h}_{t,r-1}, {\bf h}_{t-1,\widehat{R}}).
\end{align}
Note that the hidden state from the previous time step is always ${\bf h}_{t-1,\widehat{R}}$, {\em i.e.}, the one computed during the final refinement $\widehat{R}$ at time $t-1$. Our reasoning for this is that the approximation of the posterior only improves with the number of refinements~\cite{marino2018iterative}.

\noindent\textbf{Temporal Conditioning.} Inside the learning objective we set the prior and the likelihood to be conditioned on the previous frames and the refinement steps. This naturally comes from an idea that each frame is dependent on the predecessor's dynamics and therefore latent representations should follow the same property. Conditioning on the refinement steps is essential to  the iterative amortized inference procedure. To model the prior and the likelihood distributions accordingly we adopt the approach proposed in Chung \etal~\cite{chung2015recurrent} but tailor it to our iterative amortized inference setting. Specifically, the parameters of our Gaussian prior are now computed from the temporal hidden state ${\bf h}_{t-1,\widehat{R}}$:
\begin{gather}
p({\bf z}_t|{\bf x}_{<t},{\bf z}_{<t}) = \mathcal{N}({\bf z}_t; \widetilde{\mathbold{\mu}}_{t}, \text{diag}(\widetilde{\mathbold{\sigma}}_{t}^2)),\quad [\widetilde{\mathbold{\mu}}_{t}, \widetilde{\mathbold{\sigma}}_{t}] = \xi_{\theta}({\bf h}_{t-1,\widehat{R}}),
\end{gather}
where $\xi_{\theta}$ is a simple neural network with a few layers.\footnote{In practice, $\xi_{\theta}$ predicts $\log \bm\sigma_t$ for stability reasons.} Please refer to the supplemental material for details. Note that the prior only changes along the time dimension and is independent of the refinement iterations, because we refine the posterior to be as close as possible to the dynamic prior for the current time step. Finally, to complete the conditional generation, we modify the likelihood distribution as follows\footnote{Since our likelihood is a Gaussian mixture model, we are now referencing the object slot $\bullet^{(k)}$ again.}:
\begin{align}
p({\bf x}_t|{\bf x}_{<t}, {\bf z}_{\leq t, r}) = \prod_{i=1}^{D}\sum_{k=1}^{K}m_{t,r,i}^{(k)}\mathcal{N}(x_{t,i}; \mu_{t,r,i}^{(k)}, \sigma^2),\quad [m_{t,r,i}^{(k)}, \mu_{t,r,i}^{(k)}] = g_{\theta}({\bf z}_{t,r}^{(k)}, {\bf h}_{t-1,\widehat{R}}^{(k)}),
\end{align}
where $\mu_{t,r,i}^{(k)},m_{t,r,i}^{(k)}$ are mask and appearance of pixel $i$ in slot $k$ at time step $t$ and refinement step $r$. $g_{\theta}$ is a spatial mixture broadcast decoder~\cite{greff2019multi} with preceding MLP to transform the pair $\left({\bf z}_{t,r}^{(k)}, {\bf h}_{t-1,\widehat{R}}^{(k)}\right)$ into a single vector representation.
%\polina{We can define a new MLP module for combining the h and z separately.} We have added an MLP in front of the decoder
 %\polina{Andreas, I have changed the sentence about the decoder, since the modification is the MLP.}
%\polina{Generally: do we want to have more theoretical approach to the formulas? have more "mathy" derivations?  Please take a look into this paper: https://arxiv.org/pdf/1907.13052v3.pdf. It is an extension of the MoNet, very simple, but they presented it quite convincingly.}
\subsubsection{Learning and Prediction}
\vspace{-0.06in}
\label{learn}
\noindent \textbf{Architecture.} %Our architecture follows the optimization of a spatio-temporal ELBO objective (Eq.\eqref{eq:obj}) via iterative amortized inference. 
From a graphical point of view, we can think of the refinement steps and time steps as being organized on a 2D grid from Fig.~\ref{fig:overview_a}, with light gray cell $(r,t)$ representing the $r$-th refinement at time $t$. According to Eq.\eqref{eq:st-iai}, each such cell takes as input the hidden state from a previous refinement ${\bf h}_{t,{r-1}}$, the temporal hidden state ${\bf h}_{t-1,\widehat{R}}$, and the posterior parameters $\bm\lambda_{t,r}$. Outputs of each light gray cell are new posterior parameters $\bm\lambda_{t,{r+1}}$ and a new hidden state ${\bf h}_{t,r}$. At the last refinement $\widehat{R}$ at time $t$, the value of the refinement hidden state ${\bf h}_{t,r}$ is assigned to a new temporal hidden state ${\bf h}_{t,\widehat{R}}$. %Fig.~\ref{fig:overview_b} provides an insight on the internal process within one of the cells. 
%The initial values of hidden states and posterior parameters are set to zero and standard normal, respectively. Fig.~\ref{fig:overview_a} provides a high-level illustration of this view.

\noindent \textbf{Training Objective.} Instead of a direct optimization of Eq.\eqref{eq:obj}, we propose two modifications that we found to improve our model's practical performance: (1) similar to observations made by Greff~\etal~\cite{greff2019multi}, we found that color is an important factor for high-quality segmentations. In the absence of such information, we mitigate the arising ambiguity by maximizing the entropy of the masks $m_{t,r,i}^{(k)}$ along the slot dimension $k$, {\em i.e.}, we train our model by maximizing the objective
\begin{align}
\mathcal{L}_{\textrm{ELBO}} + \gamma \sum_{i=1}^{D}\sum_{k=1}^{K}m_{t,r,i}^{(k)}\log(m_{t,r,i}^{(k)}),
\label{final_obj}
\end{align}
where $\gamma$ defines the weight of the entropy loss. (2) In addition to the entropy loss, we also prioritize later refinement steps by weighting the terms in the inner sum of Eq.\eqref{eq:obj} with $\frac{r}{\widehat{R}}$.

\noindent \textbf{Prediction.}
On top of pure video decomposition, our model is also able to simulate future frames ${\bf x}_{T+1},\ldots,{\bf x}_{T+T'}$. Because our model requires image data ${\bf x}_t$ as input, which is not available during simulation of new frames, we use the reconstructed image $\widetilde{\bf x}_t$ in place of ${\bf x}_t$ to compute the likelihood $p({\bf x}_t|{\bf x}_{<t}, {\bf z}_{\leq t, r})$ in these cases. We also set the gradients $\nabla_{\bm\lambda}\mathcal{L}$, $\nabla_{\bm\mu}\mathcal{L}$, and $\nabla_{\bf m}\mathcal{L}$ to zero. %Our experimental results will show that the information carried by the temporal hidden state is powerful enough to simulate $>\!10$ frames.

\noindent \textbf{Complexity.}
Our model's ability to reuse information from previous refinements leads to a runtime complexity of $\mathcal{O}(R^2 + T)$, which is much more efficient than the $\mathcal{O}(RT)$ complexity of the traditional IODINE model~\cite{greff2019multi} (when each frame is inferred independently) in the typical case of $T\gg R$.
%\begin{algorithm}
%\caption{Training}
%\begin{multicols}{2}
%\begin{algorithmic}[1]
%\FOR{r < max(R - t, 1 )}
%\STATE $z_{t,r}^k \distas{k}  q_{\lambda}(z_{t,r}^k|x_{\leq t}, z_{< t,r}^k) $
%\STATE $\lambda_{t,r+1}^k \xleftarrow{k} \lambda_{t,r}^k + f_\phi(a^k, h_{t,r}^k, h_{t-1,\hat{R}}^k)$
%\ENDFOR
%\end{algorithmic}
%\end{multicols}
%\end{algorithm}
%\noindent \textbf{Prediction.}

%\begin{multline}
%  \mathbb{E}_{q(z_{\leq T,R}|x_{\leq T})}\Bigl[ %\sum_{t=1}^{T}\frac{1}{\hat{R}}\sum_{r=1}^{\hat{R}}(-KL(q_\lambda(z_{t,r}|x_{\leq t}, z_{< t,r}) || %p(z_t|x_{<t},z_{<t,r})) \\ + \log(p(x|x_{<t}, z_{\leq t, r})))\Bigr]
%\end{multline}
%where $\hat{R} = \max(R-t, 1)$.

%% file: experiments.tex
\section{Experiments}
\vspace{-0.1cm}
We validate our model on Bouncing Balls~\cite{van2018relational} and an augmented version of CLEVRER~\cite{yi2019clevrer}. Our experiments comprise quantitative studies of decomposition quality during generation and prediction as well as an ablation study. We complement these results with visual illustrations. An additional wide range of visualizations and experimental detailes can be found in the supplemental material. 
\vspace{-0.1cm}
\label{sec:expr}
\subsection{Setup}
\noindent \textbf{Datasets.} Bouncing Balls 
consists of $50$ frame, binary, $64\times64$ resolution video sequences.
% consists of binary video sequences with a resolution of $64\times64$ pixels and a length of 50 frames. 
Each video shows simulated balls with different masses bouncing elastically off each other and the image border. We train our model on the first 40 frames of 50K videos containing 4 balls in each frame. We use two different test sets consisting of 10K videos with 4 balls and 10K videos with 6-8 balls. We also validate our model on a color version of this \mbox{dataset that we generate using the segmentation masks.}  %, respectively. 

CLEVRER contains synthetic videos of moving and colliding objects. Each video is 5 seconds long % and contains 
(128 frames) % with a 
at resolution of $480\times 320$, which we trim and rescale to $64\times 64$ pixels (see supp. mat.).
% for details). 
For training, we use the same 10K videos as in the original source. For testing, we compute ground truth masks for the validation set using the provided annotations and test on 2.5K instances containing 3-5 objects and on 1.1K instances containing 6 objects. We set the number of slots $K$ to 6 for the CLEVRER training set and to one more than the maximum number of objects in all other cases.

%\edo{I would rewrite the paragraph above as follows (make sure that they are semantically equivalent before overwriting)\\
%For our second dataset we modify CLEVRER ~\cite{yi2019clevrer}, which contains 5 seconds (128 frames) videos of moving and colliding objects. For training, we downscale the original videos from $480\times 320$ to $64\times 64$ resolution, setting the number of slots to 6.
%For testing, we slice the videos to limit the number of static frames and compute the respective ground truth masks. We valuate our model on 2.5K videos of 3, 4 and 5 objects with 6 slots and on 1.1K videos of 6 objects with 7 slots.}

\noindent \textbf{Baselines.} We compare our approach to two recent baselines: R-NEM~\cite{van2018relational} and IODINE~\cite{greff2019multi}. R-NEM is a state-of-the-art model for unsupervised video decomposition and physics learning. While showing impressive results on simulation tasks, it is limited to binary data and has difficulties with perspective scenes. % We consider IODINE to be an appropriate model for our baseline, since our framework is built upon it. 
IODINE is more expressive but static in nature and cannot capture temporal dynamics within its probabilistic framework. However, as noted in~\cite{greff2019multi}, it can be readily applied to temporal sequences by feeding a new video frame to each iteration of the LSTM in the refinement network. We call this variant SEQ-IODINE and compare to it as well.
% With some small adjustments in the test procedure we transformed it to SEQ-IODINE and used as another baseline.

\subsection{Evaluation Metrics}
\vspace{-0.07in}
\noindent \textbf{ARI.} The Adjusted Rand Index ~\cite{rand1971objective, hubert1985comparing} is a measure of clustering similarity. It is computed by counting all pairs of samples that are assigned to the same or different clusters in the predicted and true clusterings. It ranges from -1 to 1, with score %s close to 
of $0$ indicating a random clustering and $1$ indicating a perfect match. We treat each pixel as one sample and its segmentation as the cluster assignment.

\noindent \textbf{F-ARI.} The Foreground Adjusted Rand Index is a modification of the ARI score ignoring background pixels, which often occupy the majority of the image. We argue that both metrics are necessary to assess the segmentation quality of a video decomposition method; this metric is also used in~\cite{greff2019multi, van2018relational}.
%\leon{isn't the modification here consistency with IODINE reporting procedure?} \polina{yes, it is consistent, but i wanted to explain why we are computing both.}

\noindent \textbf{MSE.} The mean squared error between pixels of the reconstructed ${\bf \widehat{x}}$ and the ground truth frames ${\bf x}$.

\subsection{Video Decomposition}
We evaluate the models on a video decomposition task at different sequence lengths.
As shown in Table~\ref{tab:DecompositionErrors} our model outperforms the baselines regardless of the presence of color information, which further reduces the error.  We are at least 7\% better than R-NEM on all metrics and at least 20\% than IODINE on ARI and MSE. Since R-NEM cannot cope well with colored data or perspective of the scenes, it is only evaluated on the Bouncing Balls dataset (binary) producing high-error results in the first frames, a phenomenon not affecting our model.
For both datasets IODINE's results are computed independently on each frame of the longest sequence, by processing frames separately IODINE does not keep the same object-slot assignment, we chose to ignore it when computing the scores.

%Table~\ref{tab:DecompositionErrors} shows the metric scores for the scene decomposition task. For the Bouncing Balls dataset we test the models on four different sequence lengths. We also test how our model and IODINE behave if the balls are colored. As can be seen from the table our method outperforms the baselines with or without color information, however, access to the color certainly improves the model performance. R-NEM shows a visible increase of the performance with the number of frames per sequence which is partially caused by a very poor results in the beginning of a sequence\footnote{We had to recompute the R-NEM ARI w/o bg results since authors used a weighted score for the first frames. }, while our model does not suffer from that issue. Since R-NEM does not work with colored data and 3D scenes, we could only compare with IODINE and SEQ-IODINE on the CLEVRER dataset. For both datasets IODINE results are computed independently for each frame on sequences of length 40. We would like to point out that by treating each frame separately IODINE does not keep the same object-slot assignment, which is a significant drawback, however we deliberately ignored it when computing the scores. 
\begin{table}[t!]
  \centering
  \caption{{\bf Quantitative Evaluation (Scene Decomposition).} We show our model's ability to produce high-quality instance segmentations for sequences with varying length. We test on sequences with 4 balls and two different types of data (binary, colored) for Bouncing Balls and on sequences with 3-5 objects for CLEVRER. Note, R-NEM does not cope with color data; hence we only run it on binary.}
%   Since R-NEM does not support non-binary data, we omit it from the corresponding experiments.}
  \resizebox{\textwidth}{!}{\begin{tabular}{ll*{12}{c}}
    \toprule
    & \multicolumn{12}{c}{{\bf Bouncing Balls}} \\
    \midrule
   & & \multicolumn{4}{c}{ARI ($\uparrow$)} & \multicolumn{4}{c}{F-ARI ($\uparrow$)} & \multicolumn{4}{c}{MSE ($\downarrow$)} \\
    \cmidrule(lr){3-6}
    \cmidrule(lr){7-10}
    \cmidrule(lr){11-14}
    \multicolumn{2}{c}{Length} & 10 & 20 & 30 & 40                                    & 10 & 20 & 30 & 40  & 10 & 20 & 30 & 40\\
    \midrule
    \multirow{4}{*}{\rotatebox[origin=c]{90}{binary}}    &R-NEM &  0.5031 & 0.6199 & 0.6632 & 0.6833       & 0.6259 & 0.7325 & 0.7708 & 0.7899      & 0.0252 & 0.0138 & 0.0096 & 0.0076 \\
        & IODINE &\multicolumn{4}{c}{0.0318}         &  \multicolumn{4}{c}{0.9986}         & \multicolumn{4}{c}{0.0018} \\
   & SEQ-IODINE &    0.0230 & 0.0223 & 0.0021  & -0.0201      & 0.8645 & 0.6028 & 0.5444 & 0.4063         & 0.0385 & 0.0782 & 0.0846 & 0.0968\\
   & Our  & {\bf0.7169} & {\bf 0.7263} & {\bf 0.7286} & {\bf 0.7294}  & {\bf 0.9999} & {\bf 0.9999} & {\bf 0.9999}  & {\bf 0.9999 } &{\bf 0.0004} & {\bf 0.0004} & {\bf 0.0004} & {\bf 0.0004}\\
    \midrule
    \multirow{3}{*}{\rotatebox[origin=c]{90}{color}}   & IODINE & \multicolumn{4}{c}{0.5841 }       & \multicolumn{4}{c}{0.9752}          & \multicolumn{4}{c}{0.0014}  \\
    & SEQ-IODINE     & 0.3789 & 0.3743 & 0.3225 & 0.2654         & 0.7517 & 0.8159 & 0.7537 & 0.6734        & 0.0160 & 0.0164 & 0.0217 & 0.0270\\
    &Our      & {\bf 0.7275} & {\bf 0.7291} & {\bf 0.7298} & \bf{0.7301}  & \bf{1.0000} & \bf{1.0000} & {\bf 0.9999} & {\bf 0.9999}  & \bf{0.0002} & \bf{0.0002} & \bf{0.0002} & \bf{0.0002}\\
    \end{tabular}}
    
    %\vspace{0.3cm}
    \resizebox{\textwidth}{!}{\begin{tabular}{ll*{12}{c}}
    \midrule
    & \multicolumn{12}{c}{{\bf CLEVRER}} \\
    \midrule
    && \multicolumn{4}{c}{ARI ($\uparrow$)} & \multicolumn{4}{c}{F-ARI ($\uparrow$)} & \multicolumn{4}{c}{MSE ($\downarrow$)} \\
    \cmidrule(lr){3-6}
    \cmidrule(lr){7-10}
    \cmidrule(lr){11-14}
   \multicolumn{2}{c}{Length}  & 10 & 20 & 30 & 40                                    & 10 & 20 & 30 & 40  & 10 & 20 & 30 & 40\\
    \midrule
   \multirow{3}{*}{\rotatebox[origin=c]{90}{color}}    & IODINE &  \multicolumn{4}{c}{0.1791} & \multicolumn{4}{c}{\bf{0.9316}}       & \multicolumn{4}{c}{0.0004}   \\
    &SEQ-IODINE    & 0.1171 & 0.1378 & 0.1558 & 0.1684 & 0.8520 & 0.8774 & 0.8780 & 0.8759       & 0.0009 & 0.0009 & 0.0010 & 0.0010 \\
     & Our   & {\bf 0.2220} & {\bf 0.2403} & {\bf 0.2555} & {\bf 0.2681}  & 0.9182 & 0.9258 & 0.9309 & 0.9312  & \bf{0.0003} & \bf{0.0003} & \bf{0.0003} & \bf{0.0003}\\
    \bottomrule
  \end{tabular}}
  \label{tab:DecompositionErrors}
  \vspace{-0.07in}
\end{table}

\begin{figure}[t]
\centering
\includegraphics[width=13.9cm]{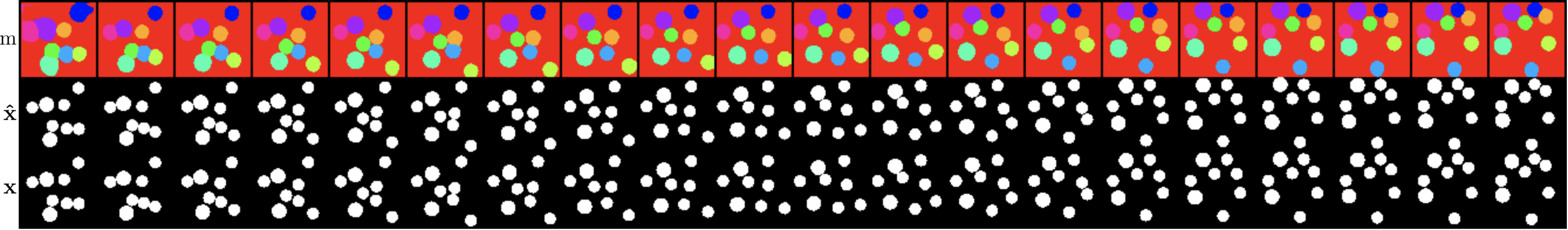} \\
\caption{{\bf Qualitative Evaluation (Bouncing Balls).} Our model can generalize to sequences with 8 balls when trained on 4 balls. Top-to-bottom: output masks, reconstructions, and ground truth video.}
\label{fig:balls_8}
\vspace{-0.07in}
\end{figure}

\subsection{Generalization}
\vspace{-0.07in}
We investigated how well our model adapts to a higher number of objects, evaluating its performance on the Bouncing Balls dataset (6 to 8 objects) and on the CLEVRER dataset (6 objects). Table~\ref{tab:generalization} shows that our F-ARI and MSE scores are at least 50\%  better than those for R-NEM, and ARI scores are just marginally worse and only on the binary data. In comparison to IODINE we are at least 4\% better across all metrics.
For the Bouncing Balls dataset we have also investigated the impact of changing the total number of possible colors to 4 and 8. The former resulting in duplicate colors for different objects and the latter in unique colors for each object. 
The higher MSE scores for the 8 balls variant is due to the model not being able to reconstruct the unseen colors. 
Sample qualitative results are shown in Fig.~\ref{fig:balls_8} and~\ref{fig:clevrer_6}, while more can be found in the supplementary material.

%In this experiment we investigated how our model can adapt to datasets with higher number of objects. We have evaluated the performance of our model on the Bouncing Balls dataset with 6 to 8 objects and on the CLEVRER dataset with 6 objects. Table~\ref{tab:generalization} shows a superior performance of our model compared to the baselines. Despite having a marginally worse ARI score compared to R-NEM, our model still significantly outperforms the baseline on ARI w/o bg and MSE. 

%For the Bouncing Ball dataset we have also investigated the effects of color on the performance. The colored training dataset had only 4 different colors of the balls. For the test data we created 2 versions: 4 and 8 colors. The former results in duplicate colors for different objects and the latter in unique colors for each object.

%We have observed that the MSE scores were significantly different for the two versions of the dataset. This difference in error is mainly caused by the model not being able to reconstruct the unseen colors, however it was still able to achieve high scores on other two metrics. The qualitative results are shown in Figures ~\ref{fig:balls_8} and~\ref{fig:clevrer_6}. For more qualitative results on all the experiments please refer to the supplemental material. 

\begin{figure}[t!]
\centering
\includegraphics[width=13.9cm]{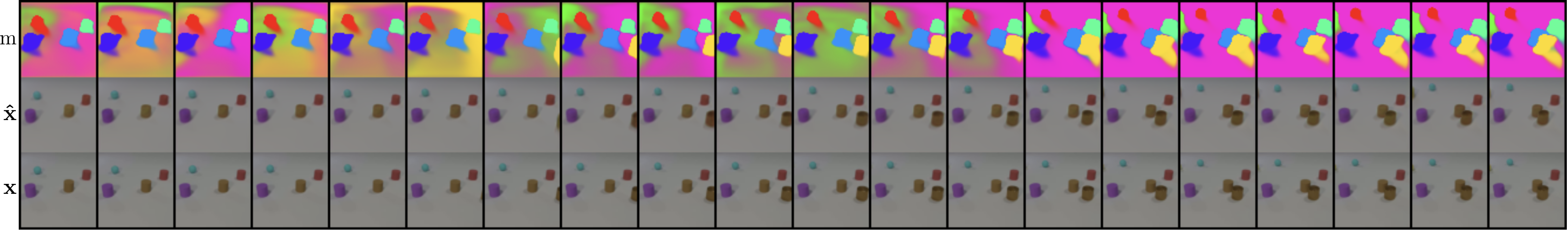} \\
\caption{{\bf Qualitative Evaluation (CLEVRER).} 
Our model can generalize to sequences with 6 objects. Furthermore, we demonstrate the ability to handle a dynamically changing number of objects, ranging from 4 in the beginning to 6 at the end.}
\label{fig:clevrer_6}
\vspace{-0.07in}
\end{figure}

\newcolumntype{P}[1]{>{\centering\arraybackslash}p{#1}}
\begin{table}[t]
    \begin{minipage}[t]{.38\textwidth}
      \centering
      \caption{{\bf Generalization}. At test time, we change the number of slots in the models from 5 to 9 for the Bouncing Balls test dataset (6-8 balls), and from 6 to 7 for the CLEVRER test dataset (6 objects).}
      \label{tab:generalization}
          \resizebox{\textwidth}{!}{\begin{tabular}{ll*{3}{c}}
            \toprule
            \multicolumn{5}{c}{{\bf Bouncing Balls}} \\
             \midrule
            && ARI ($\uparrow$) & F-ARI ($\uparrow$) & MSE ($\downarrow$)\\
            \midrule
              \multirow{4}{*}{\rotatebox[origin=c]{90}{binary}}&R-NEM &    {\bf 0.4484}         & 0.6377          & 0.0328 \\
            &     IODINE        & 0.0271         & 0.9969        & 0.0040 \\
            &  SEQ-IODINE       & 0.0263         & 0.8874         & 0.0521\\
            &    Our            & 0.4453         & {\bf 0.9999}         & {\bf 0.0008}\\
            \midrule
             \multirow{6}{*}{\rotatebox[origin=c]{90}{color}}&IODINE (4)   & 0.4136       & 0.8211        & 0.0138 \\
            &IODINE (8)   & 0.2823         &0.7197        & 0.0281 \\
            & SEQ-IODINE (4)  &0.2068    &0.5854   &0.0338 \\
            & SEQ-IODINE (8)  &  0.1571  &   0.5231   & 0.0433 \\
            &  Our (4)   &  0.4275        & {\bf 0.9998}        & {\bf 0.0004}\\
            &   Our (8)  &  {\bf 0.4317}        & 0.9900        & 0.0114\\
            \bottomrule
            \rule{0pt}{4ex} &&&&\\
            
            \toprule
            \multicolumn{5}{c}{{\bf CLEVRER}} \\
             \midrule
               & & ARI ($\uparrow$) & F-ARI ($\uparrow$) & MSE ($\downarrow$)\\
                \midrule
               \multirow{3}{*}{\rotatebox[origin=c]{90}{color}}  & IODINE&   0.2205       & 0.9305       & 0.0006\\
                & SEQ-IODINE &     0.1482   &  0.8645    & 0.0012\\
               &  Our   & {\bf 0.2839}        & {\bf 0.9355}       & {\bf 0.0004} \\
                \bottomrule
          \end{tabular}}
    \end{minipage}%
    \hfillx
    \begin{minipage}[t]{.58\textwidth}
     \caption{{\bf Ablation Study}. A 2D-LSTM extension of \mbox{IODINE} trained on sequences of 20 frames is unstable and its output segmentation lacks precision and consistency. Our efficient version of 2D-LSTM grid (Fig.~\ref{fig:overview_a}) and the conditional prior and generation increase both segmentation and reconstruction quality. Training this model on longer sequences of 40 frames we observe further improvement of the scores. Our full models including the entropy loss term~(Eq.\eqref{final_obj}) leads to the highest scores.} %This is aligned with the intuition behind the entropy term: higher entropy of the masks makes them more sparse, i.e higher level of confidence in the slot assignment.}
     \label{ablation}
       \begin{center}
       \scalebox{0.9}{ 
       \begin{tabular}[h]{@{}P{0.7cm}@{}P{0.7cm}@{}P{0.7cm}@{}P{0.7cm}@{}P{0.7cm}ccc}
            \toprule
            \rotatebox{45}{Base} & \rotatebox{45}{Grid} & \rotatebox{45}{CP+G}& \rotatebox{45}{Entropy} & \rotatebox{45}{Length} & ARI ($\uparrow$) & F-ARI ($\uparrow$) & MSE ($\downarrow$)\\
            \midrule
            \checkmark &&&& 20   & 0.0126       & 0.7765       & 0.0340\\
            \checkmark&\checkmark&\checkmark&&20 & 0.2994       & 0.9999       & 0.0010\\
            \checkmark&\checkmark&\checkmark&&40  & 0.3528       & 0.9998       & 0.0010\\
            \checkmark&\checkmark&\checkmark&\checkmark&40  &  {\bf 0.7263}        & {\bf 0.9999}      &  {\bf 0.0004} \\
            \bottomrule
          \end{tabular}}
          \end{center}
          {\small[Base: base model using 2D-LSTM; Grid: efficient triangular grid structure (Fig.~\ref{fig:overview_a}); CP+G: conditional prior and generation; Length: sequence length; Entropy: entropy term (Eq.\eqref{final_obj}]}
          
    \end{minipage} 
\end{table}

\subsection{Prediction}
\begin{figure}[h]
\begin{subfigure}{.72\textwidth}
\vspace{-0.07in}
  \centering
  % include first image
  \includegraphics[width=0.90\linewidth]{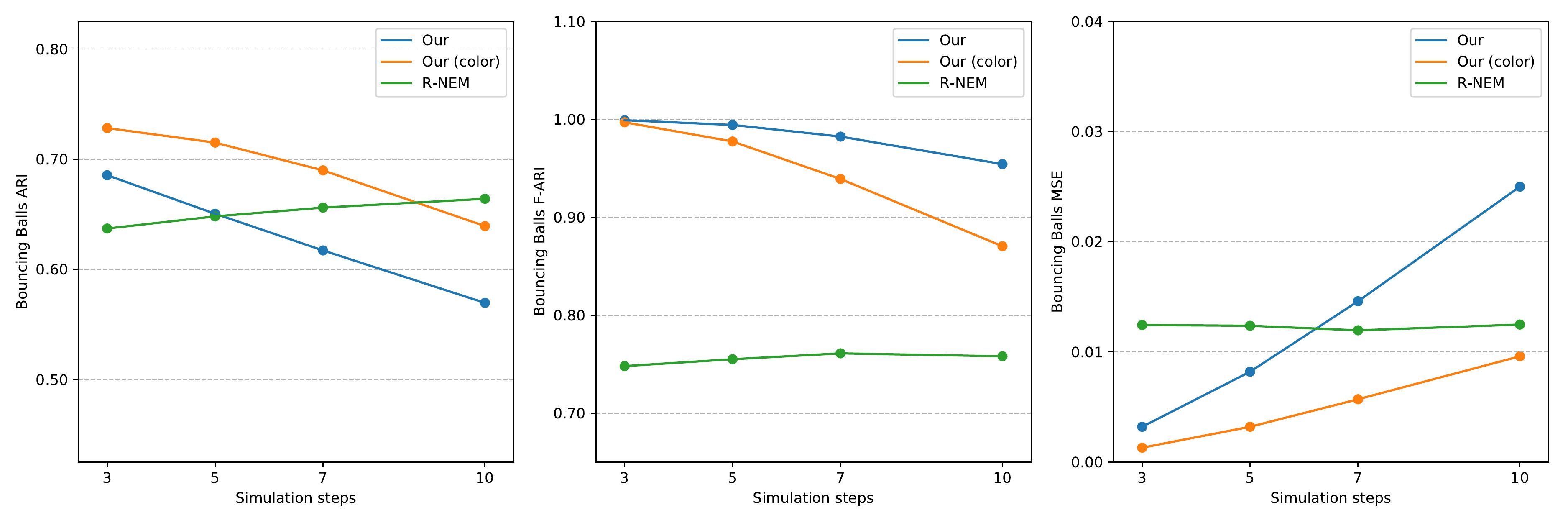}  
  \caption{Bouncing balls}
  \label{fig:sub-first}
\end{subfigure}
\begin{subfigure}{.24\textwidth}
  \centering
  % include second image
  \includegraphics[width=0.90\linewidth]{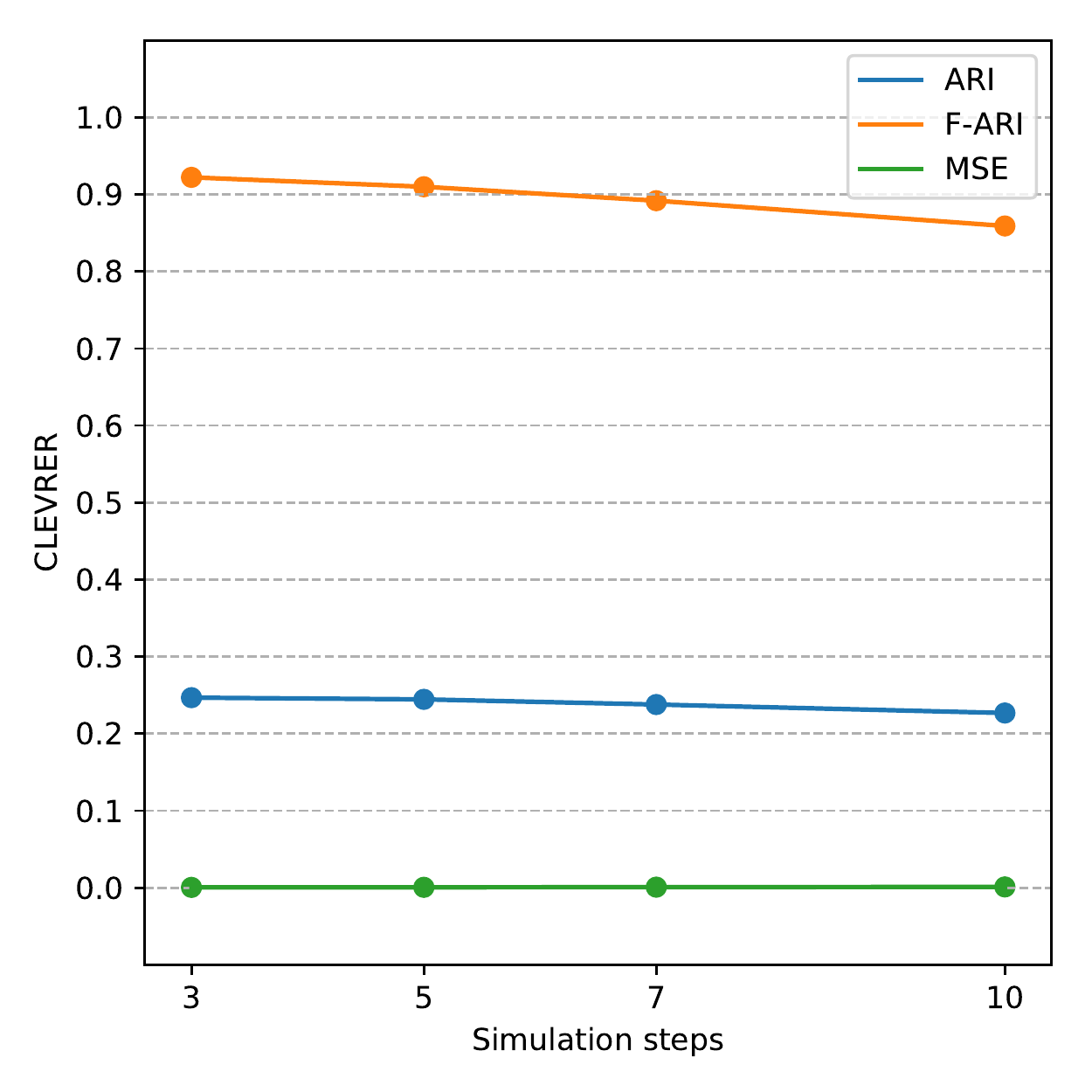}  
  \caption{CLEVRER}
  \label{fig:sub-second}
\end{subfigure}
\caption{{\bf Prediction.} We plot the prediction errors for 3, 5, 7 and 10 frames after 20 inference steps.}
\label{predict_plot}
\vspace{-0.07in}
\end{figure}

\vspace{-0.07in}
We compare the predictions of our model (Section~\ref{learn}) to those of R-NEM after 20 steps of inference on 10 predicted steps on the Bouncing Balls dataset (Fig.~\ref{predict_plot}a). As we can see from the results our model is superior to R-NEM on a shorter sequences, however for the longer sequences we are outperforming R-NEM only on colored data. We also show the prediction errors on the CLEVRER dataset in Fig.~\ref{predict_plot}b, which slowly decreases over time as expected. 

%Our model makes predictions about future objects dynamics, after several steps of learning. We run the R-NEM and our model on 20 normal steps followed by 10 prediction steps according to the prediction protocol from Section~\ref{learn}. 
%Plots from Figure~\ref{predict_plot} (a) demonstrate the prediction curves for different numbers of simulated frames on the Bouncing Balls dataset. The results show our model to be superior to R-NEM on shorter sequences, but for longer sequences we consistently outperform R-NEM only on colored data. Which signifies that color plays an important role in learning the objects dynamics in our model. 
%Plot on Figure~\ref{predict_plot} (b) shows the combined results for all scores on the CLEVRER dataset for our model. It can be seen from the plots that the quality of the predictions slightly decreases with the number of frames, however we believe this results to indicate that the model can learn and approximate the objects' dynamics.
\subsection{Ablation}
\vspace{-0.07in}
 The quantitative results for the ablation study on the binary Bouncing Balls dataset are shown in Table~\ref{ablation}. We investigate the effects of the efficient grid, conditional prior and generation, length of training sequences and entropy term on the performance of our model; all necessary and important. 

%\subsection{Disentanglement}

% Error for the bouncing balls dataset (first three tables)

%% file: conclusion.tex
\section{Conclusion}
\vspace{-0.07in}
\label{conc}
We presented a novel unsupervised learning framework capable of precise scene decomposition and dynamics modeling in multi-object videos with complex appearance and motion. The proposed approach leverages temporal consistency between latent random variables expressed through a variational energy resulting in a robust and efficient inference model. These leads to the state-of-the-art in decomposition, segmentation and prediction tasks on several datasets, one of which collected by us. Notably, our model generalizes well to more populous scenes and has improved stability in scenes with missing color information due to the entropy term.

%% file: supp.tex
\newpage
\appendix
\section*{Supplemental Material}
\label{supp}
\addcontentsline{toc}{section}{Appendices}

\renewcommand{\thesubsection}{\Alph{subsection}}
\subsection{Baselines}
\subsubsection{R-NEM}
We use the R-NEM~\cite{van2018relational} authors' original implementation and their publicly available models: \url{https://github.com/sjoerdvansteenkiste/Relational-NEM}.

\subsubsection{IODINE}
Our IODINE experiments are based on the following PyTorch implementation: \url{https://github.com/MichaelKevinKelly/IODINE}. We use the same parameters as in this code, with the exceptions of $\beta=10$ (weight factor) and, for the Bouncing Balls experiments, $R=6$ (refinement steps). The majority of the hyperparameters shared between our own model and IODINE are identical.

\subsubsection{SEQ-IODINE}
In order to test the sequential version of IODINE, we use the regularly trained IODINE model but change the number of refinement steps to the number of video frames during testing. During each refinement step, instead of computing the error between the reconstructed image and the ground truth image, we use the next video frame. Since the IODINE model was trained on $R=6$ refinement steps, extending the number of refinement steps to the video length leads to exploding gradients. This effect is especially problematic in the binary Bouncing Balls dataset with 20, 30 and 40 frames per video, because the scores of the static model are already low. We deal with this issue by clamping with max $=10$ and min $=-10$ the gradients and the $\delta$ refinement value in this experiment\footnote{ Please note that clamping was done only when applied to binary Bouncing Balls for 20, 30 and 40 frames.}. SEQ-IODINE's weak performance, especially w.r.t. the ARI, reflect the gradual divergence from the optimum as the number of frames increases.
\subsection{Datasets}
{\bf Bouncing Balls.} 
Bouncing Balls is a dataset provided by the authors of R-NEM~\cite{van2018relational}. We use the train and test splits of this dataset in two different versions: binary and color. For the color version, we randomly choose 4 colors for the 4-balls (sub-)dataset. For the 6-8 balls test data, we color them in 2 different ways: 4 colors (same as train) and 8 colors (4 from train, 4 new ones). Note that the former results in identical colors for multiple objects, while the latter guarantees unique colors for each object.

{\bf CLEVRER.} 
The version of the CLEVRER dataset~\cite{yi2019clevrer} used in this work was processed as follows:
\begin{itemize}
    \item Train split, validation split and validation annotations were obtained from the official website: \url{http://clevrer.csail.mit.edu/}. We use the validation set as test set, because the test set does not contain annotations.
    \item For training, we use the original train split. Our minimal preprocessing consists of cropping the frames along the width axis by 40 pixels on both sides, followed by a uniform downscaling to 64x64 pixels. Since the length of each video is 128 frames and the maximum number of frames during training was 40, we split the videos into multiple sequences to obtain a larger number of training samples.
    \item For testing, we trim the videos to a subsequence containing at least 3 objects and object motion. We compute these subsequences by running the script (slice\_videos\_from\_annotations.py in the attached code) from the folder with the validation split and validation annotations.
    \item The test set ground truth masks can be downloaded from \href{https://drive.google.com/file/d/1dRnBKRJXsEyKe0EaNq3SHK1KMiJOv71v/view}{\textcolor{blue}{here}}. The masks and the preprocessed test videos will be grouped into separate folders based on the number of objects in a video.
\end{itemize}

\subsection{Hyperparameters}
{\bf Initialization.}
We initialize the parameters of the posterior $\bm\lambda$ by sampling from ${\mathcal U}(-0.5,0.5)$. In all experiments, we use a latent dimensionality $\textrm{dim}(\bm z)=64$, such that $\textrm{dim}(\bm\lambda)=128$. Horizontal and vertical hidden states and cell states are of size 128, initialized with zeros. 
The variance of the likelihood is set to $\sigma=0.3$ in all experiments.

{\bf Experiments on Bouncing Balls.} For this experiment, we have explored several values of $R$ (refinement steps) and empirically found $R=6$ to be optimal in terms of accuracy and efficiency. Refining the posterior more than 6 times does not lead to any substantial improvement, however, the time and memory consumption is significantly increased. For the 4-balls dataset, we use $K=5$ slots for train and test. For our tests on 6-8 balls, we use $K=9$ slots. This protocol is identical to the one used in R-NEM~\cite{van2018relational}. Furthermore, we set $\beta = 100.0$ and scale the KL term by $\psi=10$. The weight of the entropy term is set to $\gamma=0.1$ in the binary case. As expected, the effect of the entropy term is most pronounced with binary data, so we set $\gamma=0$ in all experiments with RGB data. 

{\bf Experiments on CLEVRER.} We keep the default number of iterative refinements at $R=5$, because we did not observe any substantial improvements from a further increase. We use $K=6$ slots during training, $K=6$ slot when testing on 3-5 objects and $K=7$ slots when testing on 6 objects.

%\subsection{Architecture}

\subsection{Training}
We use ADAM~\cite{kingma2014adam} for all experiments, with a learning rate of 0.0003 and default values for all remaining parameters. During training, we gradually increase the number of frames per video, as we have found this to make the optimisation more stable. We start with sequences of length 4 and train the model until we observe a stagnant loss or posterior collapse. At the beginning of training, the batch size is 32 and is gradually decreased negatively proportional to the number of frames in the video. 

\subsection{Infrastructure and Runtime}
We train our models on 8 GeForce GTX 1080 Ti GPUs, which takes approximately one day per model.

\subsection{Discussion and Future work}
Introduction of a temporal component not only enables modelling of dynamics inside the amortized iterative inference framework but also improves the quality of the results overall. From our quantitative and qualitative comparisons with IODINE and SEQ-IODINE, we see that our model shows more accurate results on the decomposition task. We can detect new objects faster and are less sensitive to color, because our model can leverage the objects' motion cues. The ability to work with complex colored data, a property inherited from IODINE, means that we significantly outperform R-NEM. However, R-NEM is a stronger model when it comes to prediction of longer sequences, owing to its ability to model the relations between the objects in the scene. Similar ideas were used in SQAIR~\cite{kosiorek2018sequential} and GENESIS~\cite{engelcke2019genesis} by adding a relational RNN~\cite{santoro2017simple}. Integration of these concepts into our framework is a promising direction for future research. Another possible route is an application of our model to complex real-world scenarios. However, given that such datasets typically contain a much higher number of objects, as well as intricate interactions and spatially varying materials, we consider the resulting scalability questions as a separate line of research.
%--More interpretable representations
\subsection{Additional Qualitative Results}
\label{pics}
\begin{figure}[H]
\centering
\includegraphics[width=0.8\linewidth]{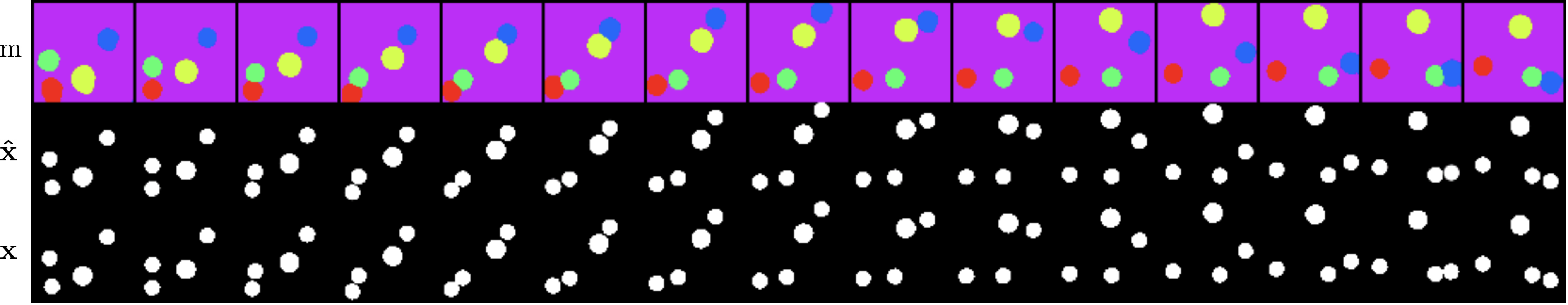}
\caption{Video decomposition using our model applied on Bouncing Balls dataset with 4 balls.}
\label{fig:bb_binary_4} 
\end{figure}

\begin{figure}[H]
\centering
\includegraphics[width=0.8\linewidth]{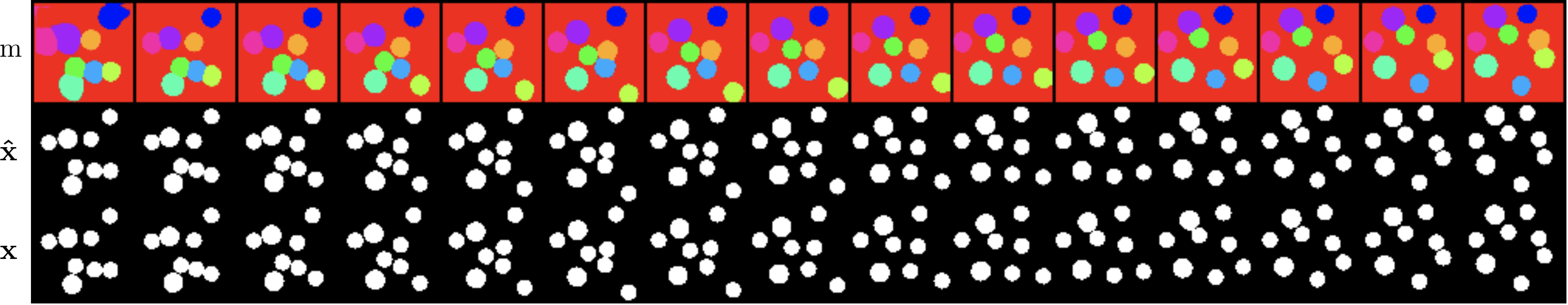}
\caption{Video decomposition using our model applied on Bouncing Balls dataset with 6-8 balls.}
\label{fig:bb_binary_6_} 
\end{figure}

\begin{figure}[H]
\centering
\includegraphics[width=1.0\linewidth]{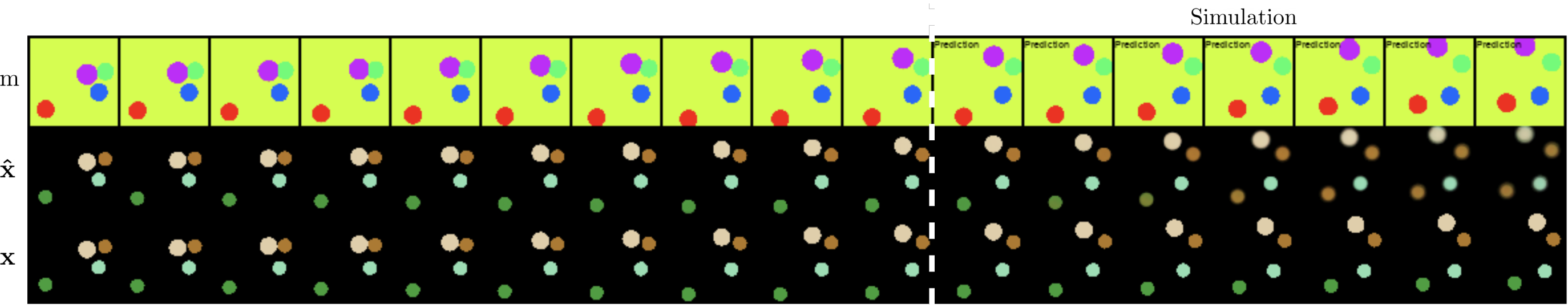}
\caption{Prediction on Bouncing Balls (colored) dataset.}
\label{fig:bb_pred} 
\end{figure}

\begin{figure}[H]
\centering
\includegraphics[width=1.0\linewidth]{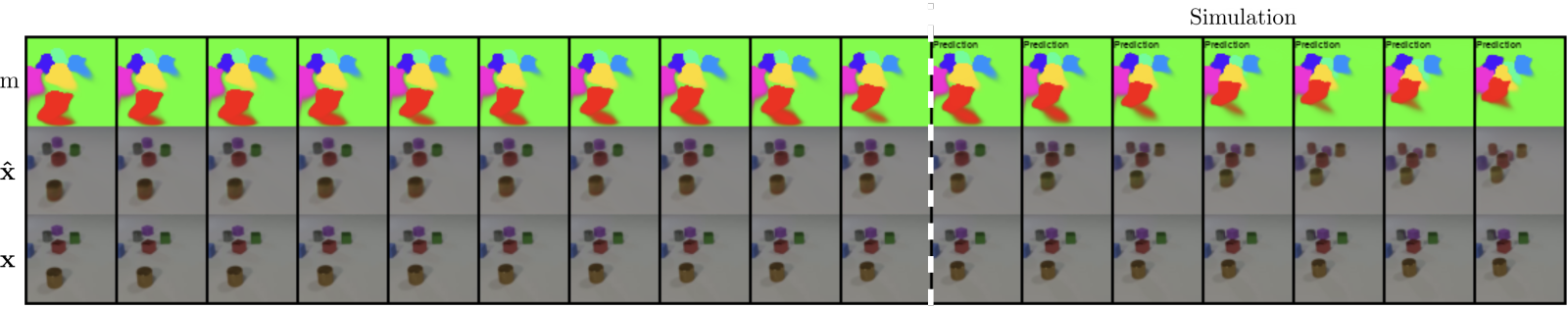}
\caption{Prediction on CLEVRER dataset. }
\label{fig:clevrer_pred} 
\end{figure}

\begin{figure}[h]
\centering
\begin{subfigure}[b]{\textwidth}
   \includegraphics[width=1\linewidth]{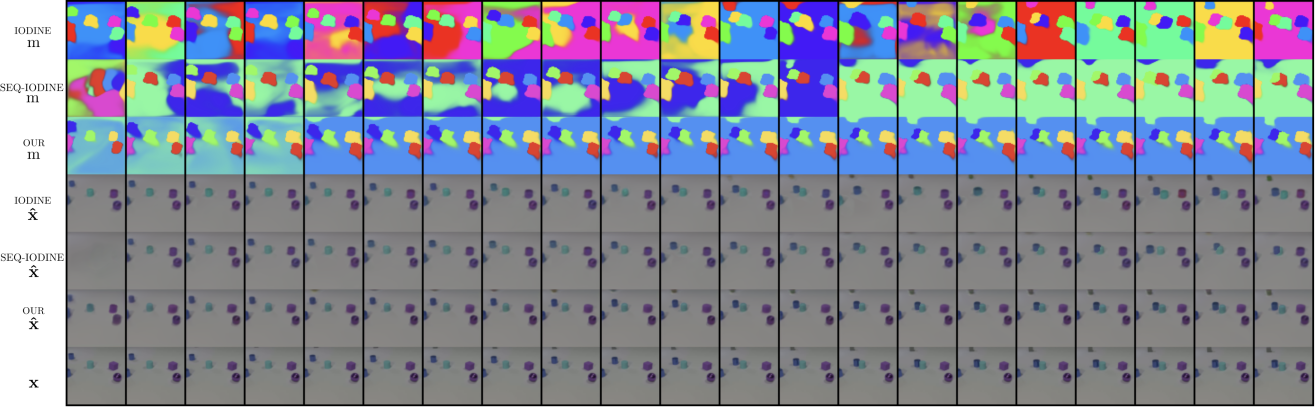}
   \caption{}
   \label{fig:iodine_vs_ours_1} 
\end{subfigure}

\begin{subfigure}[b]{\textwidth}
   \includegraphics[width=1\linewidth]{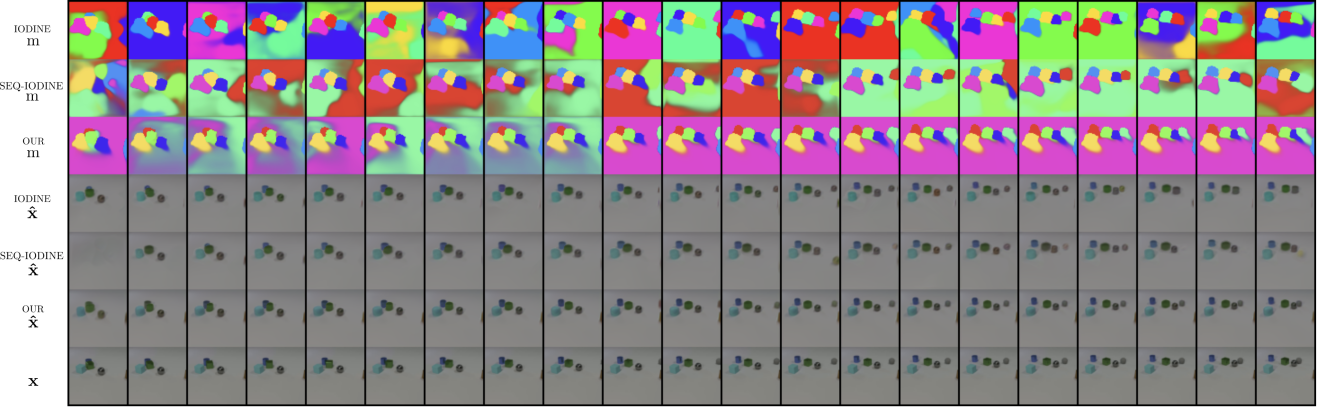}
   \caption{}
   \label{fig:iodine_vs_our_2}
\end{subfigure}

\caption[]{Qualitative results for {\bf Ours vs. IODINE vs. SEQ-IODINE} decomposition experiment. (a) From the figure it is clear that our model can much sooner detect new objects emerging to the frame, while SEQ-IODINE struggles to properly reconstruct and decompose them. And IODINE doesn't have any temporal consistence and reshuffles the slot order. (b). Here we can see that our model is much more stable with time and it does not fail to detect objects, unlike IODINE and SEQ-IODINE.}
\end{figure}
\newpage
\subsection{Disentanglement}
We demonstrate that introducing a new temporal hidden state and an additional MLP in front of the spatial broadcast decoder has not impacted its ability to separate each object's representations and disentangles them based on color, position, size and other features, similar to results shown in~\cite{greff2019multi}.
\begin{figure*}[t!]
    \centering
    \begin{subfigure}[t]{0.5237\textwidth}
        \centering
        \includegraphics[width=1.0\linewidth]{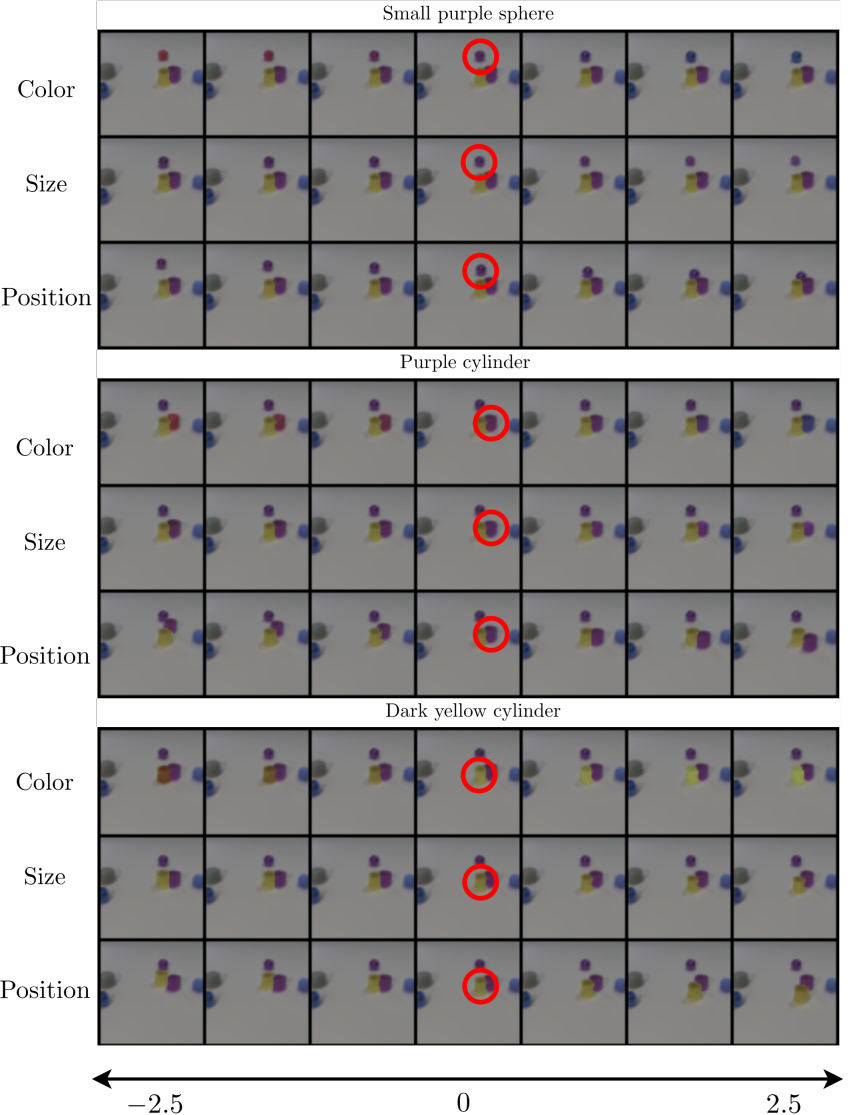}
        \caption{}
        \label{fig:latent_a}
    \end{subfigure}%
    ~
    \begin{subfigure}[t]{0.4763\textwidth}
        \centering
        \includegraphics[width=1.0\linewidth]{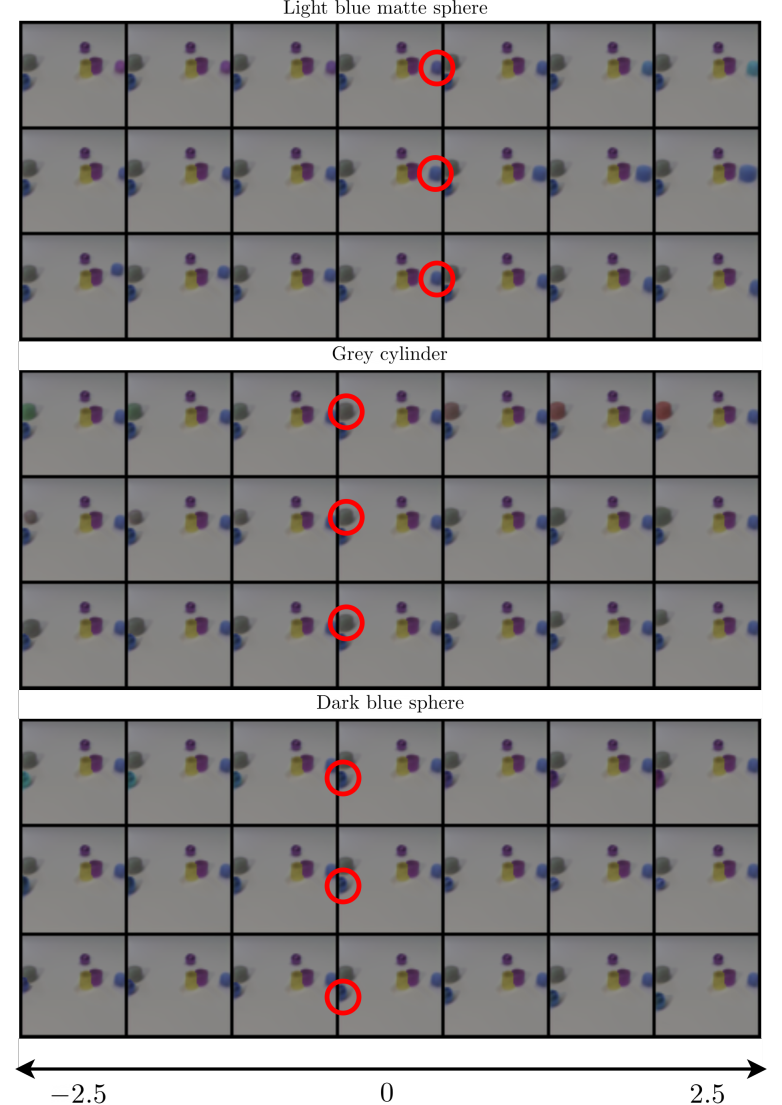}
        \caption{}
        \label{fig:latent_b}
    \end{subfigure}
    \caption{Disentanglement of the latent representations corresponding to distinct interpretable features. CLEVRER latent walks along three different dimensions: color, size and position. We chose a random frame and for each object's representation in the scene dimensions were traversed independently.}
    \label{fig:latents}
    \vspace{-0.08in}
\end{figure*}